\documentclass[sigconf]{acmart}

\usepackage{subcaption}
\AtBeginDocument{%
  \providecommand\BibTeX{{%
    \normalfont B\kern-0.5em{\scshape i\kern-0.25em b}\kern-0.8em\TeX}}}

\begin{document}
\copyrightyear{2020} 
\acmYear{2020} 
\setcopyright{acmlicensed}
\acmConference[IUI '20]{25th International Conference on Intelligent User Interfaces}{March 17--20, 2020}{Cagliari, Italy}
\acmPrice{15.00}
\acmDOI{10.1145/3377325.3377498}
\acmISBN{978-1-4503-7118-6/20/03}

\title{Proxy Tasks and Subjective Measures Can Be Misleading in Evaluating Explainable AI Systems}


\author{Zana Bu\c{c}inca*}
\affiliation{%
  \institution{Harvard University}
  \streetaddress{33 Oxford St}
  \city{Cambridge}
  \state{Massachusetts}
  \postcode{02138}
}
\email{zbucinca@seas.harvard.edu}
\author{Phoebe Lin*}
\affiliation{%
  \institution{Harvard University}
  \streetaddress{33 Oxford St}
  \city{Cambridge}
  \state{Massachusetts}
  \postcode{02138}
}
\email{phoebelin@gsd.harvard.edu}
\author{Krzysztof Z. Gajos}
\affiliation{%
  \institution{Harvard University}
  \streetaddress{33 Oxford St}
  \city{Cambridge}
  \state{Massachusetts}
  \postcode{02138}
}
\email{kgajos@eecs.harvard.edu}
\author{Elena L. Glassman}
\affiliation{%
  \institution{Harvard University}
  \streetaddress{33 Oxford St}
  \city{Cambridge}
  \state{Massachusetts}
  \postcode{02138}
}
\email{glassman@seas.harvard.edu}
\thanks{* equal contribution}


\begin{abstract}

Explainable artificially intelligent (XAI) systems form part of \em sociotechnical systems, \em e.g., human+AI teams tasked with making decisions. Yet, current XAI systems are rarely evaluated by measuring the performance of human+AI teams on actual decision-making tasks. We conducted two online experiments and one in-person think-aloud study to evaluate two currently common techniques for evaluating XAI systems: (1) using proxy, artificial tasks such as how well humans predict the AI’s decision from the given explanations, and (2) using subjective measures of trust and preference as predictors of actual performance. The results of our experiments demonstrate that evaluations with proxy tasks did not predict the results of the evaluations with the actual decision-making tasks. Further, the subjective measures on evaluations with actual decision-making tasks did not predict the objective performance on those same tasks. Our results suggest that by employing misleading evaluation methods, our field may be inadvertently slowing its progress toward developing human+AI teams that can reliably perform better than humans or AIs alone.
  
\end{abstract}



\maketitle

\section{Introduction}
Because people and AI-powered systems have complementary strengths, many expected that human+AI teams would perform better on decision-making tasks than either people or AIs alone~\cite{kamar2016directions, kamar2012combining, amershi2019guidelines}. However, there is mounting evidence that human+AI teams often perform \em worse \em than AIs alone~\cite{green2019disparate, lai2019human, poursabzi2018manipulating, green2019principles}. 

We hypothesize that this mismatch between our field's aspirations and the current reality can be attributed, in part, to several pragmatic decisions we frequently make in our research practice. Specifically, although our aspiration is formulated at the level of \em sociotechnical systems \em, i.e., human+AI teams working together to make complex decisions, we often make one of two possible critical mistakes: (1) Rather than evaluating how well the human+AI team performs together on a decision-making task, we evaluate by using proxy tasks, how accurately a human can predict the decision or decision boundaries of the AI~\cite{poursabzi2018manipulating, garcia2018explainable, lage2019human, lakkaraju2016interpretable}. (2) We rely on \em subjective measures \em of trust and preference, e.g.,~\cite{ribeiro2016should, selvaraju2017grad, weitz2019you}, instead of objective measures of performance. We consider each of these two concerns in turn.

First, evaluations that use proxy tasks force study participants to pay attention to the AI and the accompanying explanations---something that they are unlikely to do when performing a realistic decision-making task.
Cognitive science provides compelling evidence that people treat cognition like any other form of labor~\cite{kool18:mental} and favor less demanding forms of cognition, i.e., heuristics over analytical thinking, even in high stakes contexts like medical diagnosis~\cite{lambe16:dual}. Therefore, we hypothesize that user performance and preference on proxy tasks may not accurately predict their performance and preference on the actual decision-making tasks where their cognitive focus is elsewhere and they can choose whether and how much to attend to the AI.

Second, subjective measures such as trust and preference have been embraced as the focal point for the evaluation of explainable systems~\cite{ribeiro2016should, selvaraju2017grad, weitz2019you}, but we hypothesize that subjective measures may also be poor predictors of the ultimate performance of people performing realistic decision-making tasks while supported by explainable AI-powered systems. Preference and trust are important facets of explainable AI systems: they may predict users' intent to attend to the AI and its explanations in realistic tasks settings and adhere to the system’s recommendations. However, the goal of explainable interfaces should be instilling in users the \emph{right} amount of trust~\cite{dzindolet2003role, muir1987trust, lee2004trust}. This remains a remarkable challenge, as on one end of the trust spectrum users might over-rely on the system and remain oblivious of its errors, whereas on the other end they might exhibit self-reliance and ignore the system’s correct recommendations. 
Furthermore, evaluating an AI's decision, its explanation of that decision, and incorporating that information into the decision-making process requires cognitive effort and the existing evidence suggests that preference does not predict performance on cognitive tasks~\cite{deslauriers2019measuring,shah2008heuristics,garbarino1997cognitive}.

To evaluate these two hypotheses, we conducted two online experiments and one in-person study of an AI-powered decision support system for a nutrition-related decision-making task. In one online study we used a proxy task, in which participants were asked to predict the AI's recommendations given the explanations produced by the explainable AI system. In the second online study, participants completed an actual decision-making task: actually making decisions assisted by the same explainable AI system as in the first study. In both studies, we measured participants' objective performance and collected subjective measures of trust, preference, mental demand, and understanding. In the in-person study, we used a think-aloud method to gain insights into how people reason while making decisions assisted by an explainable AI system. In each study, we presented participants with two substantially distinct explanation types eliciting either deductive or inductive reasoning.

The results of these studies indicate that (1) subjective measures from the proxy task do not generalize to the actual decision-making task, and (2) when using actual decision-making tasks, subjective results do not predict objective performance results.
Specifically, participants trusted and preferred inductive explanations in the proxy task, whereas they trusted and preferred the deductive explanations in the actual task. Second, in the actual decision-making task, participants recognized AI errors better with inductive explanations, yet they preferred and trusted the deductive explanations more. The in-person think-aloud study revealed insights about why participants preferred and trusted one explanation type over another, but we found that by thinking aloud during an actual decision-making task, participants may be induced to exert additional cognitive effort, and behave differently than they would
during an actual decision-making task when they are, more realistically, not thinking aloud.

In summary, we show that the results of evaluating explainable AI systems using proxy tasks may not predict the results of evaluations using actual decision-making tasks. Users also do not necessarily perform better with systems that they prefer and trust more. To draw correct conclusions from empirical studies, explainable AI researchers should be wary of evaluation pitfalls, such as proxy tasks and subjective measures. Thus, as we recognize that explainable AI technology forms part of sociotechnical systems, and as we increasingly use these technologies in high-stakes scenarios, our evaluation methodologies need to reliably demonstrate how the entire sociotechnical systems (i.e., human+AI teams) will perform on real tasks.

\section{Related Work}

\subsection{Decision-making and Decision Support Systems}

Decision-making is a fundamental cognitive process that allows humans to choose one option or course of action from among a set of alternatives~\cite{wilson2001encyclopedia, wang2007theoretical, wang2006layered}.
Since it is an undertaking that requires cognitive effort, people often employ mental shortcuts, or heuristics, when making decisions~\cite{tversky1974judgment}.
These heuristics save time and effort, and frequently lead to good outcomes, but in some situations they result in cognitive biases that systematically lead to poor decisions (see, e.g.,~\cite{blumenthal2015cognitive}). 

To help people make good decision reliably, computer-based Decision Support Systems (DSS) have been used across numerous disciplines (e.g., management~\cite{gorry1971framework}, medicine~\cite{johnston1994effects}, justice~\cite{zeleznikow2004building}). While DSS have been around for a long time, they are now increasingly being deployed because the recent advancements in AI enabled these systems to achieve high accuracy. But since humans are the final arbiters in decisions made with DSS, the overall sociotechincal system's accuracy depends both on the system's accuracy and on the humans and their underlying cognitive processes. Research shows that even when supported by a DSS, people are prone to insert bias into the decision-making process~\cite{green2019disparate}.

One approach for mitigating cognitive biases in decision-making is to use cognitive forcing strategies, which introduce self-awareness and self-monitoring of decision-making~\cite{croskerry2003cognitive}. Although not universally effective~\cite{sherbino14:ineffectiveness}, these strategies have shown promising results as they improve decision-making performance, both if the human is assisted~\cite{green2019principles, poursabzi2018manipulating} or is not assisted by a DSS~\cite{lambe16:dual}. To illustrate, Green \& Chen~\cite{green2019principles} showed that across different AI-assisted decision-making treatments, humans performed best when they had to make the preliminary decision on their own first before being shown the system recommendation (which forced them to engage analytically with the system's recommendation and explanation if their own preliminary decision differed from that offered by the system). 
Even though conceptual frameworks that consider cognitive processes in decision-making with DSS have been proposed recently~\cite{wang2019designing}, further research is needed to thoroughly investigate how to incorporate DSS into human decision-making and the effect of cognitive processes while making system-assisted decisions.

\subsection{Evaluating AI-Powered Decision Support Systems}

Motivated by the growing number of studies in interpretable and explainable AI-powered decision support systems, researchers have called for more rigorous evaluation of explainable systems~\cite{doshi2017towards, hoffman2018metrics, gilpin2018explaining}. Notably, Doshi-Velez \& Kim~\cite{doshi2017towards} proposed a taxonomy for evaluation of explainable AI systems, composed of the following categories: application grounded evaluation (i.e., domain experts evaluated on actual tasks), human grounded evaluation (i.e., lay humans evaluated on simplified tasks) and functionally grounded evaluation (i.e., no humans, proxy tasks). To put our work into context, our definition of \emph{the actual task} falls into application grounded evaluation, where people for whom the system is intended (i.e., not necessarily experts) are evaluated on the intended task. Whereas, the \emph{the proxy task} is closer to human grounded evaluation but addresses both domain experts and lay people evaluated on simplified tasks, such as the simulation of model's prediction given an input and an explanation. 

Studies using actual tasks evaluate the performance of human and the system, as a whole, on the decision-making task~\cite{kleinberg2017predictions, yin2019understanding, green2019principles, bansal2019beyond}. In these studies, participants are told to focus on making good decisions and it is up to them to decide whether and how to use the AI's assistance to accomplish the task.
In contrast, studies that use proxy tasks evaluate how well users are able to simulate the model's decisions \cite{poursabzi2018manipulating, garcia2018explainable, lage2019human, chang09:reading} or decision boundaries \cite{lakkaraju2016interpretable}. In such studies, participants are specifically instructed to pay attention to the AI. These studies evaluate the human's mental model of the system when the human is actively attending to the system's predictions and explanations, but do not necessarily evaluate how well the human is able to perform real decision-making tasks with the system. For example, to identify which factors make a model more interpretable, Lage et al. ask participants to simulate the interpretable model's predictions~\cite{lage2019human}.

In addition to the evaluation task, the choice of evaluation metrics is a critical one for the correct evaluation of intelligent systems~\cite{arnold20:predictcive}. In explainable AI literature, subjective measures, such as user trust and experience, have been largely embraced as the focal point for the evaluation of explainable systems~\cite{ribeiro2016should, selvaraju2017grad, weitz2019you, Zhou2018InterpretableBD}. Hoffman et al.~\cite{hoffman2018metrics} proposed metrics for explainable systems that are grounded in the subjective evaluation of a system (e.g., user satisfaction, trust, and understanding). These may take the form of questionnaires on attitude and confidence in the system~\cite{hauslschmid2017supportingtrust} and helpfulness of the system~\cite{kulesza2015principles, cai2019human}. However, while these measures are informative, evidence suggests they do not necessarily predict user's performance with the system. For example, Green \& Chen~\cite{green2019disparate} discovered that self-reported measures could be misleading, since participant's confidence in their performance was negatively associated with their actual performance. Similarly, Lai \& Tan~\cite{lai2019human} found that humans cannot accurately estimate their own performance. More closely related to our findings, Poursabzi-Sangdeh et al.~\cite{poursabzi2018manipulating} observed that even though participants were significantly more confident on the predictions of one model over the other, their decisions did not reflect the stated confidence. Furthermore, Lakkaraju \& Bastani~\cite{lakkaraju2019fool} demonstrated that participants trusted the same underlying biased model almost 10 times more when they were presented with misleading explanations compared to the truthful explanations that revealed the model's bias. These findings indicate that not only are subjective measures poor predictors of performance, but they can easily be manipulated and lead users to adhere to biased or malicious systems.

\section{Experiments}

We conducted experiments with two different evaluation tasks and explanation designs to test the following hypotheses:\\
\textbf{H1:} Results of widely accepted proxy tasks, where the user is asked to explicitly engage with the explanations, may not predict the results of realistic settings where the user's focus is on the actual decision-making task.\\
\textbf{H2:} Subjective measures, such as self-reported trust and preference with respect to different explanation designs, may not predict the ultimate human+AI performance.

\subsection{Proxy Task}

\begin{figure*}[t]
\begin{subfigure}{0.48\linewidth}
\includegraphics[width=\linewidth]{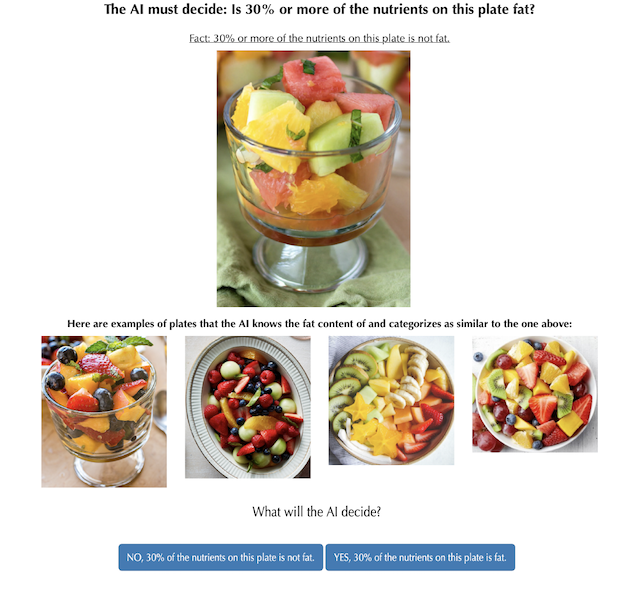}
\caption{} \label{fig:1a}
\end{subfigure}
\hspace*{\fill} 
\begin{subfigure}{0.48\linewidth}
\includegraphics[width=\linewidth]{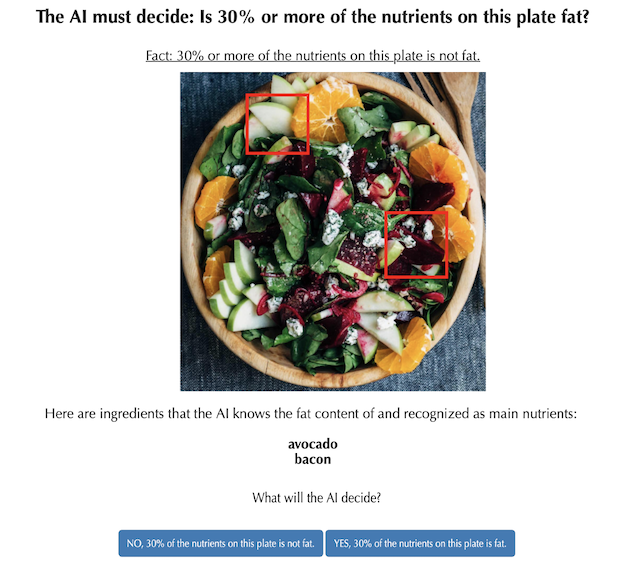}
\caption{} \label{fig:1b}
\end{subfigure}
\caption{The proxy task. Illustration of the simulated AI system participants interacted with: (a) is an example of an inductive explanation with appropriate examples. (b) is an example of a deductive explanation with misrecognized ingredients, where the simulated AI misrecognized apples and beets as avocados and bacon.} \label{fig:1}
\end{figure*}

\subsubsection{Task Description}
We designed the task around nutrition because it is generally accessible and plausibly useful in explainable AI applications for a general audience. Participants were shown a series of 24 images of different plates of food. The ground truth of the percent fat content was also shown to them as a fact. Participants were then asked: ``What will the AI decide?'' given that the AI must decide ``Is X\% or more of the nutrients on this plate fat?''. As illustrated in Figure~\ref{fig:1}, each image was accompanied by explanations generated by the simulated AI. The participants chose which decision they thought the AI would make given the explanations and the ground truth. 

We designed two types of explanations, eliciting either inductive or deductive reasoning. In inductive reasoning, one infers general patters from specific observations. Thus, for the inductive explanations, we created example-based explanations that required participants to recognize the ingredients that contributed to fat content and draw their own conclusion about the given image. As shown in Figures~\ref{fig:1a}, the inductive explanations began with ``Here are examples of plates that the AI knows the fat content of and categorizes as similar to the one above.'' Participants then saw four additional images of plates of food.
In deductive reasoning, in contrast, one starts with general rules and reaches a conclusion with respect to a specific situation. Thus, for the deductive explanations, we provided the general rules that the simulated AI applied to generate its recommendations. For example, in Figure \ref{fig:1b}, the deductive explanation begins with ``Here are ingredients the AI knows the fat content of and recognized as main nutrients:'' followed by a list of ingredients.

We chose a within-subjects study design, where for one half of the study session, participants saw inductive explanations and, for the other half of the study session, they saw deductive explanations. The order in which the two types of explanations were seen was counterbalanced. Each AI had an overall accuracy of 75\%, which meant that in 25\% of the cases the simulated AI misclassified the image or misrecognized ingredients (e.g., Figure \ref{fig:1b}). The order of the specific food images was randomized, but all participants encountered the AI errors at the same positions. We fixed the errors at questions 4, 7, 11, 16, 22 and 23, though which food the error was associated to was randomized. We included the ground truth of the fat content of plates of food, because the main aim of the proxy task was to measure whether the user builds correct mental models of the AI and not to assess the actual nutrition expertise of the participant. 

\subsubsection{Procedure}
This study was conducted online, using Amazon Mechanical Turk. Participants were first presented with brief information about the study and an informed consent form. Next, participants completed the main part of the study, in which they answered 24 nutrition-related questions, divided into two blocks of 12 questions. They saw inductive explanations in one block and the deductive explanations in the other. The order of explanations was randomized across participants. 
Participants completed mid-study and end-study questionnaires so that they would provide a separate assessment for each of the two explanation types. They were also asked to directly compare their experiences with the two simulated AIs in a questionnaire at the end of the study.

\subsubsection{Participants}
We recruited 200 participants via Amazon Mechanical Turk (AMT). Participation was limited to adults in the US. Of the total 200 participants, 183 were retained for final analyses, while 17 were excluded based on their answers to two common-sense questions included in the questionnaires (i.e., \textit{What color is the sky?}). The study lasted 7 minutes on average. Each worker was paid 2 USD. 

\subsubsection{Design and Analysis}
This was a within-subjects design. The within-subjects factor was explanation type --- inductive or deductive.

We collected the following measures: 

\begin{itemize}
    \item Performance: Percentage of correct predictions of AI's decisions 
    \item Appropriateness: Participants responded to the statement \textit{``The AI based its decision on appropriate examples/ingredients.''} with either 0=No or 1=Yes (after every question)
    \item Trust: Participants responded to the statement \textit{``I trust this AI to assess the fat content of food.''} on a 5-point Likert scale from 1=Strongly disagree to 5=Strongly agree (at the end of each block) 
    \item Mental demand: Participants answered the question  \textit{``How mentally demanding was understanding how this AI makes decisions?''} on a 5-point Likert scale from 1=Very low to 5=Very high (every four questions)
    \item Comparison between the two explanation types: Participants were asked at the end of the study to choose one AI over another on trust, preference, and mental demand. 
\end{itemize}

We used repeated measures ANOVA for within-subjects analyses and the binomial test for the comparison questions.

\subsection{Actual Decision-making Task}

\begin{figure*}[!h]
\begin{subfigure}{0.48\linewidth}
\includegraphics[width=\linewidth]{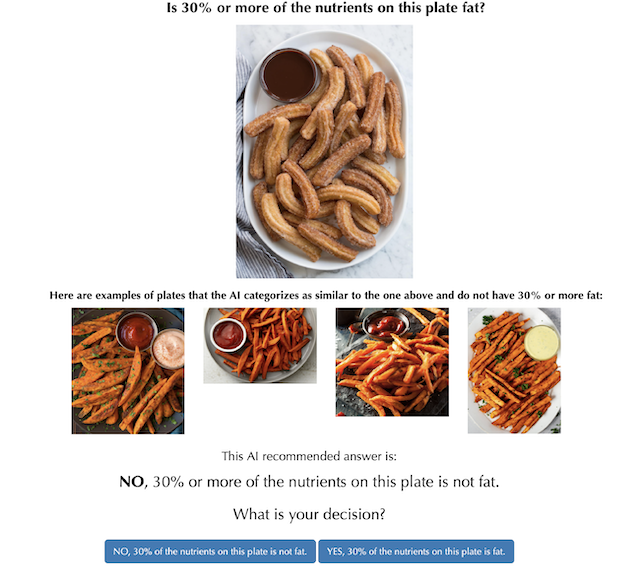}
\caption{} \label{fig:2a}
\end{subfigure}
\hspace*{\fill} 
\begin{subfigure}{0.48\linewidth}
\includegraphics[width=\linewidth]{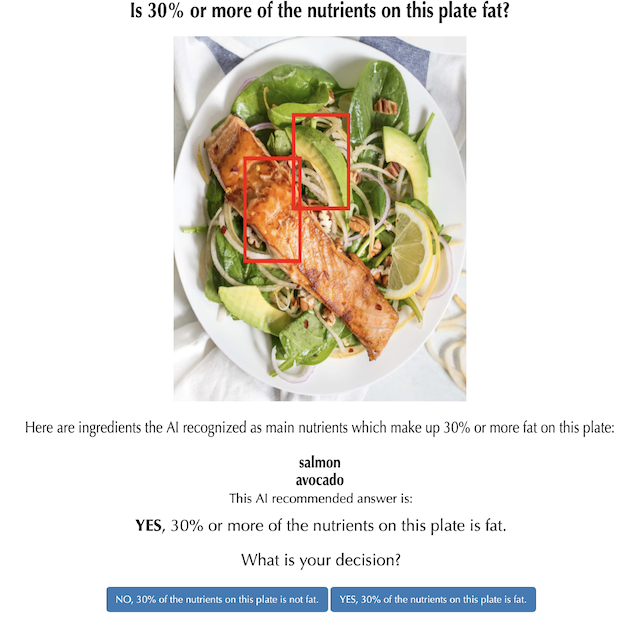}
\caption{} \label{fig:2b}
\end{subfigure}
\caption{The actual task. Illustration of the simulated AI system participants interacted with. (a) is an example of incorrect recommendations with inductive explanations. Contrasting the query image with the explanations reveals that the simulated AI misrecognized churros with chocolate as sweet potato fries with BBQ sauce. (b) is an example of correct recommendation with deductive explanations.} \label{fig:2}
\end{figure*}

\subsubsection{Task description}

\begin{figure}[!h]
\begin{subfigure}{0.48\linewidth}
\includegraphics[width=\linewidth]{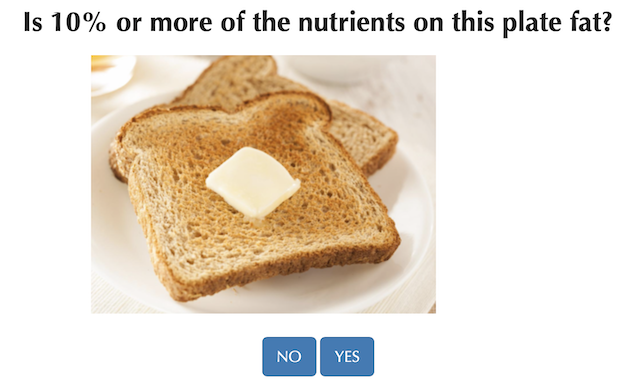}
\caption{} \label{fig:noAI}
\end{subfigure}
\hspace*{\fill} 
\begin{subfigure}{0.48\linewidth}
\includegraphics[width=\linewidth]{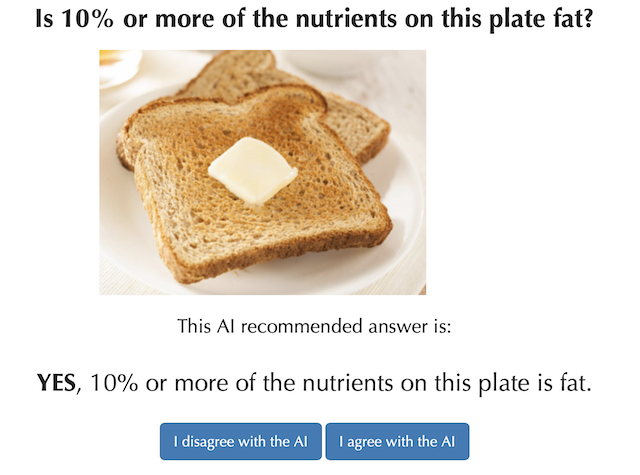}
\caption{} \label{fig:noExp}
\end{subfigure}
\caption{The baseline conditions. (a) no AI (b) no explanations} \label{fig:baselines}
\end{figure}

The actual decision-making task had a similar set up to the proxy task. Participants were shown the same series of 24 images of different plates of food, but were asked their own decision whether the percent fat content of nutrients on the plate is higher than a certain percentage.  As illustrated in Figure~\ref{fig:2}, each image was accompanied by an answer recommended by a simulated AI, and an explanation provided by that AI. 
We introduced two more conditions to serve as baselines in the actual decision-making task depicted in Figure \ref{fig:baselines}.

There were three between-subjects conditions in this study: 1. the no-AI baseline (where no recommendations or explanations were provided), 2. the no-explanation baseline (where a recommendation was provided by a simulated AI, but no explanation was given), and 3. the main condition in which both recommendations and explanations were provided. In this last condition, two within-subjects sub-conditions were present: for one half of the study participants saw inductive explanations and for the other they saw deductive explanations. The order in which the two types of explanations were seen was counterbalanced. In the no-AI baseline, participants were not asked any of the questions relating to the performance of the AI.

The explanations in this task differed only slightly from the explanations in the proxy task, because they indicated the AI's recommendation. Inductive explanations started with: ``Here are examples of plates that the AI categorizes as similar to the one above and do (not) have X\% or more fat.'' followed by four examples of images. Similarly, deductive explanations stated: ``Here are ingredients the AI recognized as main nutrients which do (not) make up X\% or more fat on this plate:'' followed by a list of ingredients.

\subsubsection{Procedure}
The procedure was the same as for the proxy task. The study was conducted online, using the Amazon Mechanical Turk. Participants were first presented with a brief information about the study and an informed consent form. Next, participants completed the main part of the study, in which they answered 24 nutrition-related questions, divided into two blocks of 12 questions.

All participants also completed a questionnaire at the end of the study, providing subjective assessments of the system they interacted with. Participants who were presented with AI-generated recommendations accompanied by explanations also completed a mid-study questionnaire (so that they would provide separate assessment for each of the two explanation types) and they were also asked to directly compare their experiences with the two simulated AIs at the end of the study.

\subsubsection{Participants}
We recruited 113 participants via Amazon Mechanical Turk (AMT). Participation was limited to adults in the US. Of the total 113 participants, 102 were retained for final analyses, while 11 were excluded based on their answers to two common-sense questions included in the pre-activity and post-activity questionnaires (i.e., \textit{``What color is the sky?''}). The task lasted 10 minutes on average. Each worker was paid 5 USD per task. 

\subsubsection{Design and Analysis}
This was a mixed between- and within-subjects design. As stated before, the three between-subjects conditions were: 1. the no-AI baseline; 2. the no-explanation baseline, in which the AI-generated recommendations were provided but no explanations; 3. the main condition, in which both the AI-generated recommendations and explanations were provided. The within-subjects factor was explanation type (inductive or deductive) and it was applied only for participants who were presented with AI-generated recommendations with explanations.   

We collected the following measures: 

\begin{itemize}
    \item Performance: Percentage of correct answers (overall for each AI, and specifically for questions when AI presented incorrect explanations)
    \item Understanding: Participants responded to the statement \textit{``I understand how the AI made this recommendation.''} on a 5-point Likert scale from 1=Strongly disagree to 5=Strongly agree (after every question)
    \item Trust: Participants responded to the statement \textit{``I trust this AI to assess the fat content of food.''} on a 5-point Likert scale from 1=Strongly disagree to 5=Strongly agree (every four questions) 
    \item Helpfulness: Participants responded to the statement \textit{``This AI helped me assess the percent fat content.''} on a 5-point Likert scale from 1=Strongly disagree to 5=Strongly agree (at the end of each block)
    \item Comparison between the two explanation types: Participants were asked at the end of the study to choose one AI over another on trust, preference, understanding and helpfulness. 
\end{itemize}

We used analysis of variance (ANOVA) for between-subjects analyses and repeated measures ANOVA for within-subjects analyses. We used the binomial test for the comparison questions.

\section{Results}

\subsection{Proxy Task Results}

The explanation type had a significant effect on participants' trust and preference in the system. Participants trusted the AI more when presented with inductive explanations ($M=3.55$), rather than deductive explanations ($M=3.40$, $F_{1,182}=5.37, p = .02$). Asked to compare the two AIs, most of the participants stated they trusted more the inductive AI ($58\%, p= .04$). When asked the hypothetical question: \textit{``If you were asked to evaluate fat content of plates of food, which AI would you prefer to interact with more?''}, again most of the participants ($62\%$) chose the inductive AI over the deductive AI ($p= .001$). 

The inductive AI was also rated significantly higher ($M=0.83$) than the deductive AI ($M=0.79$) in terms of the appropriateness of examples (ingredients for the deductive condition) on which the AI based its decision ($F(1,182) = 13.68, p=0.0003$). When the AI presented incorrect examples/ingredients, there was no significant difference among the inductive ($M=0.47$) and deductive ($M=0.50$) conditions ($F(1,182) = 1.02, p= .31, n.s.$).

We observed no significant difference in overall performance when participants were presented with inductive ($M=0.64$) or deductive explanations ($M=0.64$, $F(1,182) = 0.0009, n.s.$). When either AI presented incorrect explanations, although the average performance dropped for both inductive ($M=0.40$) and deductive ($M=0.41$) conditions, there was also no significant difference among them ($F(1,182) = .03, n.s.$).

In terms of mental demand, there was a significant effect of the explanation type. Participants rated the deductive AI ($M=2.94$) as more mentally demanding than the inductive AI ($M=2.79$, $F(1,182) = 7.75, p = .0006$). The effect was noticed also when they were asked: \textit{``Which AI required more thinking while choosing which decision it would make?''}, with $61 \%$ of participants choosing deductive over inductive $(p=.005)$.

\begin{figure*}[!h]
\begin{subfigure}{0.48\linewidth}
\includegraphics[width=\linewidth]{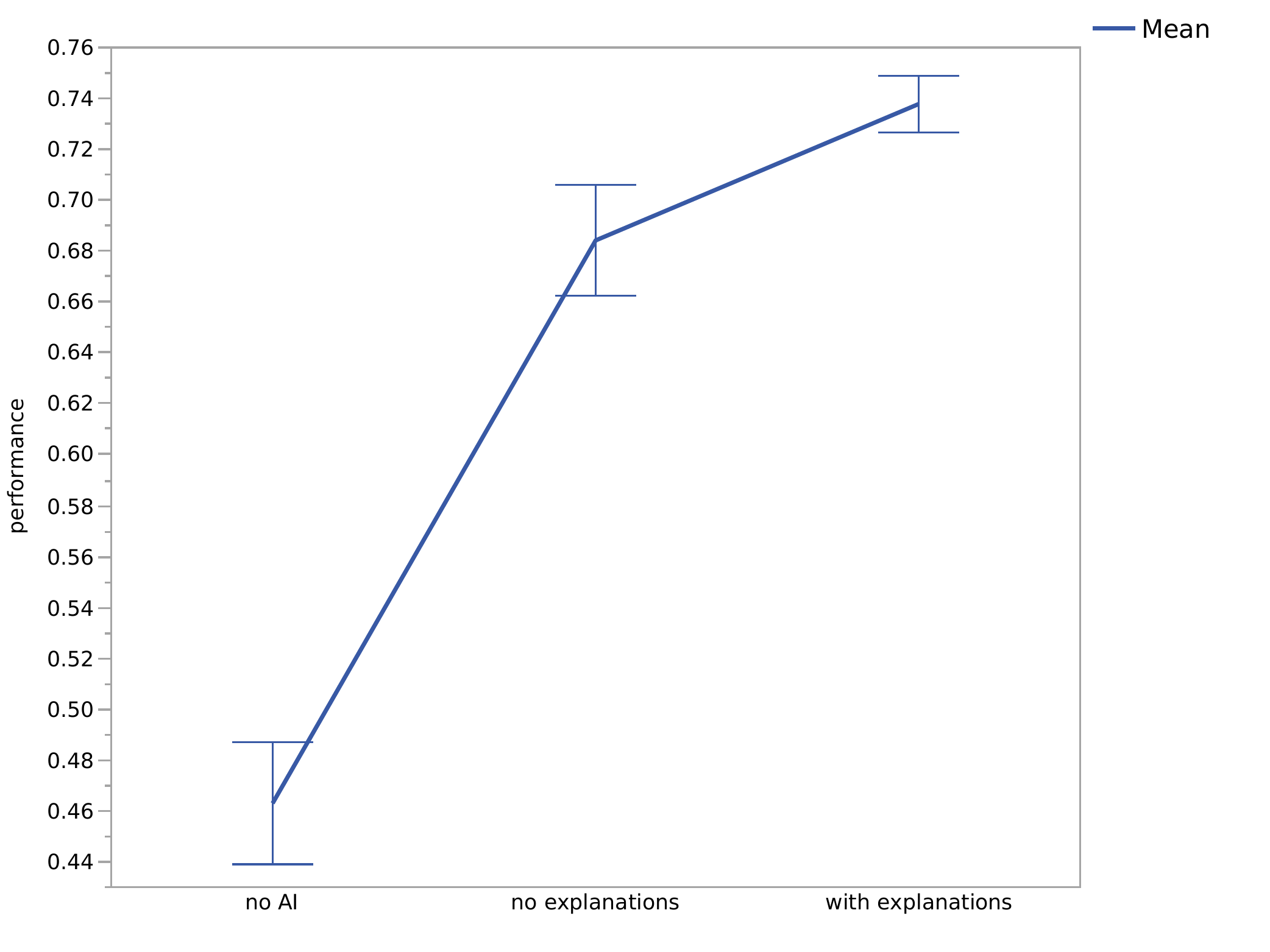}
\caption{} \label{fig:e2e_baseline}
\end{subfigure}
\hspace*{\fill} 
\begin{subfigure}{0.48\linewidth}
\vspace*{0.1in}
\includegraphics[scale=0.42]{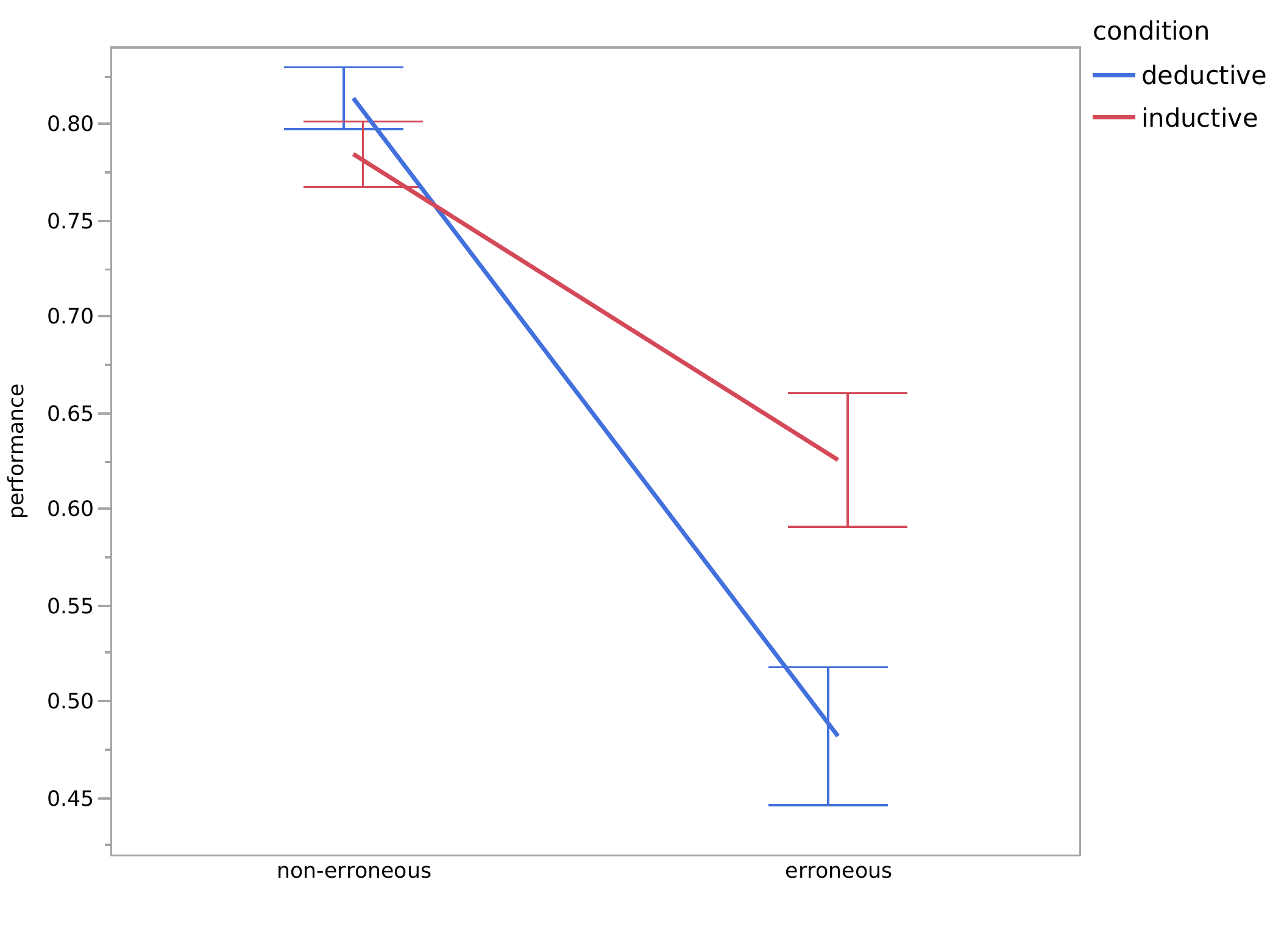}
\caption{} \label{fig:e2e_err}
\end{subfigure}
\caption{Performance in the actual decision-making task. (a) depicts the mean of performance among no-AI, no-Explanations and with-Explanations (overall) conditions.  (b) depicts the mean of performance among inductive and deductive conditions, when the AI recommendation is correct and erroneous. Error bars indicate one standard error.} \label{fig:e2e}
\end{figure*}

\subsection{Actual Decision-making Task Results}

18 participants were randomized into the no-AI condition, 19 into the AI with no explanation condition, and 65 were presented with AI recommendations supported by explanations.

We observed a significant main effect of the presence of explanations on participants' trust in the AI's ability to assess the fat content of food. Participants who saw either kind of explanation, trusted the AI more ($M=3.56$) than those who received AI recommendations, but no explanations ($M=3.17$, $F_{1,483} = 11.28, p = .0008$). Further, there was a significant main effect of the explanation type on participants' trust: participants trusted the AI when they received deductive explanations more ($M=3.68$) than when they received inductive explanations ($M=3.44$, $F_{1,64}=5.96, p = .01$). When asked which of the two AIs they trusted more, most participants ($65\%$) said that they trusted the AI that provided deductive explanations more than the one that provided inductive explanations ($p=.02$). 

Participants also found the AI significantly more helpful when explanations were present ($M=3.78$) than when no explanations were offered ($M=3.26$, $F_{1, 147} = 4.88, p = .03$). Further, participants reported that they found deductive explanations more helpful ($M=3.92$) than inductive ones ($M=3.65$) and this difference was marginally significant ($F_{1,64}=3.66, p = .06$). When asked which of the two AIs they found more helpful, most participants ($68\%$) chose the AI that provided deductive explanations ($p=.006$).

Participants also reported that they understood how the AI made its recommendations better when explanations were present ($M=3.84$) than when no explanations were provided ($M=3.67$, $F_{1,2014} = 6.89 p = .009$). There was no difference in the perceived level of understanding between the two explanation types ($F_{1,64} = 0.44, p = .51$).

Asked about their overall preference, most participants ($63\%$) preferred the AI that provided deductive explanations over the AI that provided inductive explanations ($p=.05$).

In terms of actual performance on the task, participants who received AI recommendations (with or without explanations) provided a significantly larger fraction of accurate answers ($M=0.72$) than those who did not receive AI recommendations ($M=0.46$, $F_{1,2446} = 118.07, p < .0001$). Explanations further improved overall performance: participants who saw explanations of AI recommendations had a significantly higher proportion of correct answers ($M=0.74$) than participants who did not receive explanations of AI recommendations ($M=0.68$, $F_{1,2014} = 5.10, p = .02$) (depicted in Figure \ref{fig:e2e_baseline}). There was no significant difference between the two explanation types in terms of overall performance ($F_{1,64} = 0.44, n.s.$). However, we observed a significant interaction between explanation type and the correctness of AI recommendations ($F_{2,2013} = 15.03 p < .0001$). When the AI made correct recommendations, participants performed similarly when they saw inductive ($M=.78$) and deductive ($M=.81$) explanations ($F_{1,64} = 1.13, n.s.$). When the AI made incorrect recommendations, however, participants were significantly more accurate when they saw inductive ($M=0.63$) than deductive ($M=0.48$) explanation ($F_{1,64} = 7.02, p = .01$) (depicted in Figure \ref{fig:e2e_err}).

To ensure the results of our studies were not random, we replicated both experiments with almost identical setup and obtained the same main results (in terms of significance) reported in this section.

\begin{figure}[!h]
\includegraphics[width=\linewidth]{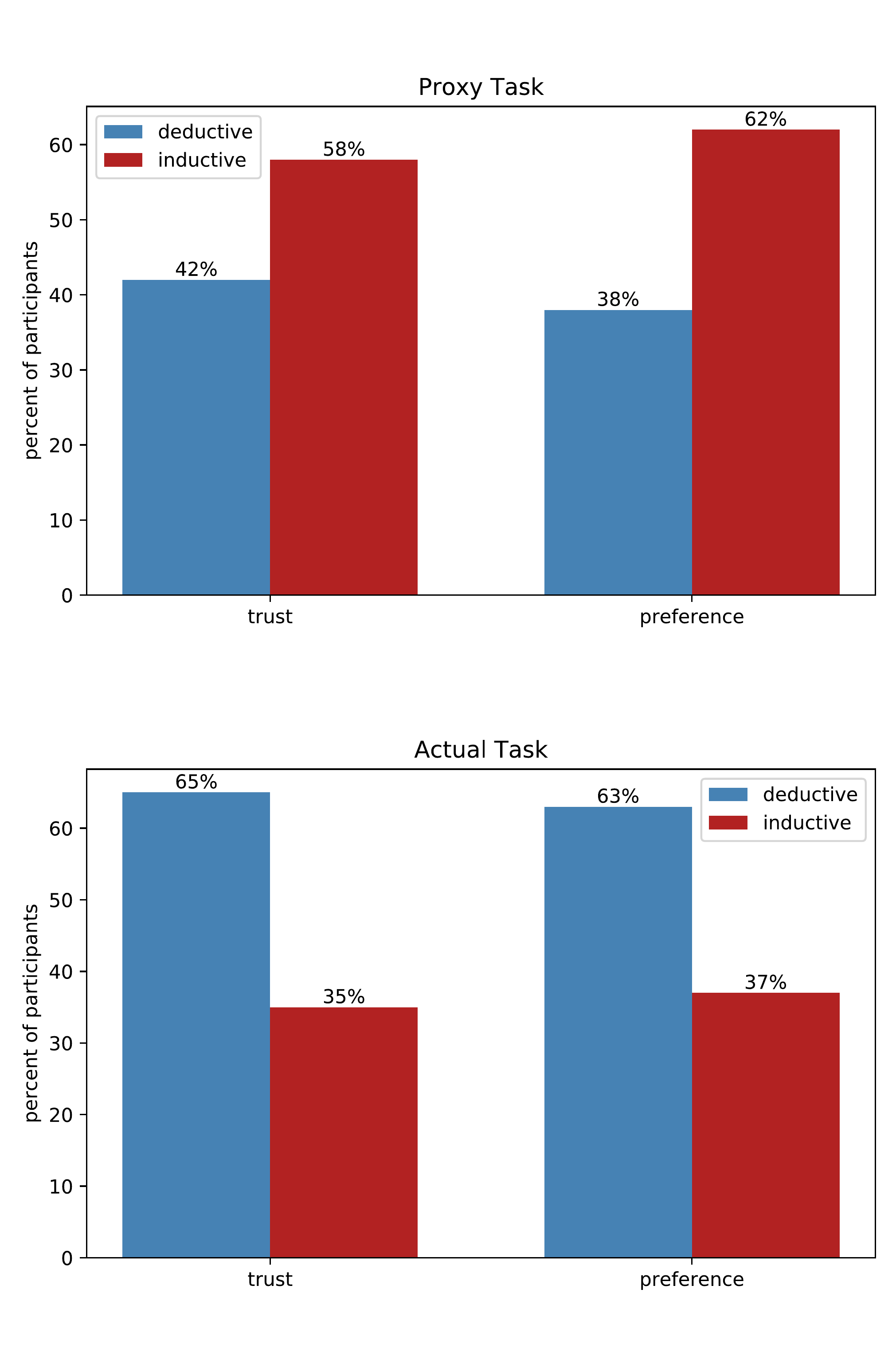}
\caption{Subjective evaluations in terms of trust and preference of the two AIs. Red and blue depict the percent of participants that chose inductive and deductive AI, respectively. (a) proxy task (b) actual decision-making task.} \label{fig:subjective}
\end{figure}

\section{Qualitative Study}
Through the qualitative study, we explored the user reasoning and sought to gain insight into the discrepancy between subjective measures and performance. We asked participants to think aloud during an in-person study in order to understand how and why people perceive AI the way they do, in addition to what factors go into making decisions when assisted by an AI.

\subsection{Task}
The same task design was used in this study as in the actual decision-making task, except that all participants were presented with the main condition (where both recommendations and explanations were provided). As in the actual decision-making task, each participant saw both inductive and deductive explanations.

\subsection{Procedure}
Upon arriving to the lab, participants were presented with an informed consent form, including agreeing to being screen- and audio-recorded, and instructions on the task. Afterwards, the steps in this study were similar to those in the actual decision-making task, except that we added the think-aloud method~\cite{ericsson84:protocol}: as participants completed the task, they were asked to verbalize their thought process as they made each decision. At the end of the task, there was a semi-structured interview, during which participants briefly discussed how they believed the two AIs were making their recommendations and why they did or did not trust them. Participants also discussed if and why they preferred one AI over the other.

\subsection{Participants}
We recruited 11 participants via community-wide emailing lists (8 female, 3 male, age range 23--29, M = 24.86, SD = 2.11). Participants were primarily graduate students with backgrounds from design, biomedical engineering, and education. Participants had varying levels of experience with AI and machine learning, ranging from 0--5 years of experience.  

\subsection{Design and Analysis}

We transcribed the think-aloud comments and the post-task interviews.
Transcripts were coded and analyzed for patterns using an inductive approach~\cite{inductive-analysis}. We focused on comments about (1) how the AI made its recommendations; (2) trust in the AI; (3) erroneous recommendations; (4) why people preferred one explanation type over the other.  From a careful reading of the transcripts, we discuss some of the themes and trends that emerged from the data. 

\subsection{Results}
\textbf{Preference of one explanation type over another.} 
Eight out of the 11 participants preferred the inductive explanations. Participants who preferred inductive explanations perceived the four images as data. One participant stated that \textit{``Because [the AI] showed similar pictures, I knew that it had data backing it up''} (P3). On the other hand, participants who preferred deductive explanations perceived the listing of ingredients to be reliable, and that \textit{``if the AI recognized that it's steak, then I would think, Oh the AI knows more about steak fat than I do, so I'm going to trust that since it identified the object correctly.''} (P6). 

In our observations, we found that the way participants used the explanations was different depending on the explanation type. With inductive explanations, one participant often first made their own judgement before looking at the recommendation, and then used the recommendation to confirm their own judgement. In a cake example, one participant said, \textit{``So I feel it probably does have more than 30\% because it's cake, and that's cream cheese. But these are all similar to that, and  the AI also says that it does have more than 30\% fat, so I agree''} (P2). With deductive explanations, participants evaluated the explanations and recommendation more before making any decision. In the same cake example, a different participant said, \textit{``There are [the AI recognizes] nuts, cream cheese, and cake. That seems to make sense. Nuts are high in fat, so is dairy, so I agree with that.''} (P6).

\textbf{Cognitive Demand.}
At the end of the study, participants were asked which AI was easier to understand. Ten out of 11 participants felt the inductive explanations were easier to understand than the deductive explanations. Several participants stated that the deductive explanations forced them to think more, and that generally they spent more time making a decision with deductive explanations. One participant said, for example, \textit{``I feel like with this one I have to think a bit more and and rely on my own experiences with food to see or understand to gauge what's fatty.''} (P2).

\textbf{Errors and Over-reliance.}
Nine out of 11 participants claimed to trust the inductive explanations more. We intentionally introduced erroneous recommendations because we expected participants to utilize them to calibrate their mental model of the AI. When participants understood the error and believed the error was reasonable for an AI to make, they expressed less distrust in subsequent questions. However, when participants perceived the error to be inconsistent with other errors, their trust in subsequent recommendations was hurt much more. For example, one participant stated, \textit{``I think the AI makes the recommendation based on shape and color. But in some other dessert examples, it was able to identify the dessert as a dessert. So I wasn't sure why it was so difficult to understand this particular item''} (P5). 

We found that there was also some observable correlation between explanation type and trust. Many participants claimed it was easier to identify errors from the inductive explanations, yet agreed with erroneous recommendations from inductive explanations more. In some of those instances, participants either did not realize the main food image was different from the other four or felt the main food image was similar enough though not exact. Lastly, one participant stated the inductive explanations were easier to understand because \textit{``you can visually see exactly why it would come to its decision,''}, but for deductive explanations \textit{``you can see what it's detecting but not why''} (P8), and yet this participant also stated that the deductive explanations seemed more trustworthy.

\textbf{Impact of the Think-Aloud method on participant behavior.}
In this study, we asked participants to perform the actual decision-making task and we expected to observe similar results to those obtained in the previous experiment when using the actual tasks.
Yet, in this study, 8 out of the 11 participants preferred the inductive explanations and 10 out of 11 participants felt the inductive explanations were easier to understand than the deductive explanations. These results are comparable to the results we obtained in the previous experiment when we used the proxy task rather than the actual task. 

We believe that the use of the think-aloud method may have impacted participants' behavior in this study. Specifically, because participants were instructed to verbalize their thoughts, they were more likely to engage in analytical thinking when considering the AI recommendations and explanations than they were in the previous experiment with the actual tasks, where their focus was primarily on making decisions.

It is possible that while the think-aloud method is part of standard research practice for evaluating interfaces, it is itself a form of cognitive forcing intervention~\cite{croskerry2003cognitive}, which impacts how people perform on cognitively-demanding tasks such as interacting with an explainable AI system on decision-making tasks. The act of talking about the explanations led participants to devote more of their attention and cognition to the explanations, and thus made them behave more similarly to participants in working with the proxy task rather than those working with the actual task.

\section{Discussion}

In this study, we investigated two hypotheses regarding the evaluation of AI-powered explainable systems: 

\begin{itemize}
\item \textbf{H1:} Results of widely accepted proxy tasks, where the user is asked to explicitly engage with the explanations, may not predict the results of realistic settings where the user's focus is on the actual decision-making task.
\item \textbf{H2:} Subjective measures, such as self-reported trust and preference with respect to different explanation designs, may not predict the ultimate human+AI performance. 
\end{itemize}
We examined these hypotheses in the context of a nutrition-related decision-making task, by designing two distinct evaluation tasks and two distinct explanation designs. The first task was a proxy task, where the users had to simulate the AI's decision by examining the explanations. The second task was the more realistic, actual decision-making task, where the user had to make their own decisions about the nutritional content of meals assisted by AI-generated recommendations and explanations. Each of the tasks had two parts, where users interacted with substantially different explanation styles---inductive and deductive. 

In the experiment with the proxy task, participants preferred and trusted the AI that used inductive explanations significantly more. They also reported that the AI that used inductive explanations based its decision on more accurate examples on average than the AI that used deductive explanations. When asked \textit{``If you were asked to evaluate fat content of plates of food, which AI would you prefer to interact with more?''}), the majority of participants chose the AI that provided inductive explanations. 

In contrast with the proxy task experiment, in the experiment with the actual decision-making task, participants rated the AI with deductive explanations as their preferred AI, and viewed it as more trustworthy and more helpful compared to the AI that used inductive explanations. 

The contrast in terms of performance measures was less pronounced. When attempting proxy tasks, participants demonstrated nearly identical accuracy regardless of explanation type. However, when attempting actual decision-making tasks and the AI provided an incorrect recommendation, participants ignored that incorrect recommendation and provided the correct answer significantly more often when they had access to inductive, not deductive, explanations for the AI's recommendation.

These contradictory results produced by the two experiments indicate that results of evaluations that use proxy tasks may not correspond to results on actual tasks, thus supporting \textbf{H1}. This may be because in the proxy task the users cannot complete the task without engaging analytically with the explanations. Whereas, in the actual decision-making task, the user's primary goal is to make the most accurate decisions about the nutritional content of meals; she chooses whether and how deeply she engages with the AI's recommendations and explanations. 

This finding has implications for the explainable AI community, as there is a current trend to use proxy tasks to evaluate user mental models of the AI-powered systems, with the implicit assumption that the results will translate to the realistic settings where users make decisions about an actual task while assisted by an AI. 

We tested \textbf{H2} on the actual decision-making task.
The results show that participants preferred, trusted and found the AI with deductive explanations more helpful than the AI that used inductive explanations. Yet, they performed significantly better with the AI that used inductive explanations when the AI made erroneous recommendations. Therefore, \textbf{H2} is also supported. This finding suggests that the design decisions for explainable interfaces should not be made by relying solely on user experience and subjective measures. Subjective measures of trust and preference are, of course, valuable and informative, but they should be used to \em complement \em rather than replace performance measures.

Our research demonstrated that results from studies that use proxy tasks may not predict results from studies that use realistic tasks. Our results also demonstrated that user preference may not predict their performance. However, we recognize that evaluating novel AI advances through human subjects experiments that involve realistic tasks is expensive in terms of time and resources, and may negatively impact the pace of innovation in the field. Therefore, future research needs to uncover \emph{why} these differences exist so that we can develop low burden evaluation techniques that correctly predict the outcomes of deploying a system in a realistic setting.

We believe that the reason why explainable AI systems are sensitive to the difference between proxy task and actual task evaluation designs is that different AI explanation strategies require different kinds and amounts of cognition from the users (like our inductive and deductive explanations). However, people are reluctant to exert cognitive effort~\cite{kool2010decision,kool18:mental} unless they are motivated or forced to do so. They also make substantially different decisions depending on whether they choose to exert cognitive effort or not~\cite{garbarino1997cognitive,shah2008heuristics}. In actual decision-making situations, people often choose \em not \em to engage in effortful analytical thinking, even in high-stakes situations like medical diagnosis~\cite{lambe16:dual}. Meanwhile, proxy tasks force participants to explicitly pay attention to the behavior of the AI and the explanations produced. Thus, results observed when participants interact with proxy tasks do not accurately predict people's behavior in many realistic settings. In our study, participants who interacted with the proxy task \emph{felt} that the deductive explanations required significantly more thinking than the inductive explanations. Therefore, in the proxy task where the participants were obliged to exert cognitive effort to evaluate the explanations, they said they preferred and trusted the less cognitively demanding explanations more, the inductive explanations. In contrast, in the actual task the participants could complete the task even without engaging with the explanations. Thus, we suspect that in the deductive condition participants perceived the explanations as too mentally demanding, and chose to over-rely on the AI's recommendation, just to avoid cognitive effort of examining those explanations. They also might have perceived the AI that provided deductive explanations as more competent \emph{because} it required more thinking.

One implication of our analysis is that the effectiveness of explainable AI systems can be substantially impacted by the design of the interaction (rather than just the algorithms or explanations). For example, a recent study showed that a simple cognitive forcing strategy (having participants make their own preliminary decision before being shown the AI's decision) resulted in much higher accuracy of the final decisions made by human+AI teams than any strategy that did not involve cognitive forcing~\cite{green2019principles}. 

Inadvertently, we uncovered an additional potential pitfall for evaluating explainable AI systems. As the results of our qualitative study demonstrated, the use of the think-aloud method---a standard technique for evaluating interactive systems---can also substantially impact how participants allocate their mental effort.
Because participants were asked to think aloud, we suspect that they exerted additional cognitive effort to engage with the explanations and analyze their reasoning behind their decisions. 

Together, these results indicate that cognitive effort is an important aspect of explanation design and its evaluation. Explanations high in cognitive demand might be ignored by the users while simple explanations might not convey the appropriate amount of evidence that is needed to make informed decisions. At the same time, traditional methods of probing users' minds while using explainable interfaces should also be re-evaluated. By taking into account the cognitive effort and cognitive processes that are employed during the evaluation of the explanations, we might generate explainable interfaces that optimize the performance of the sociotechnical (human+AI) system as a whole. Such interfaces would instill trust, and make the user aware of the system's errors.

\section{Conclusion}
To achieve the aspiration of human+AI teams that complement one-another and perform better than either the human or the AI alone, researchers need to be cautious about their pragmatic decisions. In this study, through online experiments and an in-person study, we showed how several assumptions researchers make about the evaluation of the explainable AI systems for decision-making tasks could lead to misleading results. 

First, choosing proxy tasks for the evaluation of explainable AI systems shifts the user's focus toward the AI, so the obtained results might not correspond to results of the user completing the actual decision-making task while assisted by the AI. In fact, our results indicate that users trust and prefer one explanation design (i.e. inductive) more in the proxy task, while they trust and prefer the other explanation design (i.e. deductive) more in the actual decision-making task.

Second, the subjective evaluation of explainable systems with measures such as trust and preference may not correspond to the ultimate user performance with the system. We found that people trusted and preferred the AI with deductive explanations more, but recognized AI errors better with the inductive explanations. 

Lastly, our results suggest that think-aloud studies may not convey how people make decisions with explainable systems in realistic settings. The results from the think-aloud in-person study, which used the actual task design, aligned more with the results we obtained in the proxy task. 

These findings suggest that to draw correct conclusions about their experiments, explainable AI researchers should be wary of the explainable systems' evaluation pitfalls and design their evaluation accordingly. Particularly, the correct and holistic evaluation of explainable AI interfaces as sociotechnical systems is of paramount importance, as they are increasingly being deployed in critical decision-making domains with grave repercussions.

\vspace{3mm}\noindent\textbf{Acknowledgements.}
We would like to thank Tianyi Zhang and Isaac Lage for helpful feedback.

\bibliography{iui2020}


\begin{thebibliography}{00}


\ifx \showCODEN    \undefined \def \showCODEN     #1{\unskip}     \fi
\ifx \showDOI      \undefined \def \showDOI       #1{{\tt DOI:}\penalty0{#1}\ }
  \fi
\ifx \showISBNx    \undefined \def \showISBNx     #1{\unskip}     \fi
\ifx \showISBNxiii \undefined \def \showISBNxiii  #1{\unskip}     \fi
\ifx \showISSN     \undefined \def \showISSN      #1{\unskip}     \fi
\ifx \showLCCN     \undefined \def \showLCCN      #1{\unskip}     \fi
\ifx \shownote     \undefined \def \shownote      #1{#1}          \fi
\ifx \showarticletitle \undefined \def \showarticletitle #1{#1}   \fi
\ifx \showURL      \undefined \def \showURL       #1{#1}          \fi

\bibitem{amershi2019guidelines}
{Saleema Amershi}, {Dan Weld}, {Mihaela Vorvoreanu}, {Adam Fourney}, {Besmira
  Nushi}, {Penny Collisson}, {Jina Suh}, {Shamsi Iqbal}, {Paul~N Bennett},
  {Kori Inkpen}, {and} {others}. 2019.
\newblock \showarticletitle{Guidelines for human-AI interaction}. In {\em
  Proceedings of the 2019 CHI Conference on Human Factors in Computing
  Systems}. ACM, 3.
\newblock


\bibitem{arnold20:predictcive}
{Kenneth~C. Arnold}, {Krysta Chaunce}, {and} {Krzysztof~Z. Gajos}. 2020.
\newblock \showarticletitle{Predictive Text Encourages Predictable Writing}. In
  {\em Proceedings of the 25th International Conference on Intelligent User
  Interfaces} {\em (IUI '20)}. ACM, New York, NY, USA.
\newblock


\bibitem{bansal2019beyond}
{Gagan Bansal}, {Besmira Nushi}, {Ece Kamar}, {Walter~S Lasecki}, {Daniel~S
  Weld}, {and} {Eric Horvitz}. 2019.
\newblock \showarticletitle{Beyond Accuracy: The Role of Mental Models in
  Human-AI Team Performance}. In {\em Proceedings of the AAAI Conference on
  Human Computation and Crowdsourcing}, Vol.~7. 2--11.
\newblock


\bibitem{blumenthal2015cognitive}
{Jennifer~S Blumenthal-Barby} {and} {Heather Krieger}. 2015.
\newblock \showarticletitle{Cognitive biases and heuristics in medical decision
  making: a critical review using a systematic search strategy}.
\newblock {\em Medical Decision Making\/} {35}, 4 (2015), 539--557.
\newblock


\bibitem{cai2019human}
{Carrie~J Cai}, {Emily Reif}, {Narayan Hegde}, {Jason Hipp}, {Been Kim},
  {Daniel Smilkov}, {Martin Wattenberg}, {Fernanda Viegas}, {Greg~S Corrado},
  {Martin~C Stumpe}, {and} {others}. 2019.
\newblock \showarticletitle{Human-centered tools for coping with imperfect
  algorithms during medical decision-making}. In {\em Proceedings of the 2019
  CHI Conference on Human Factors in Computing Systems}. ACM, 4.
\newblock


\bibitem{chang09:reading}
{Jonathan Chang}, {Sean Gerrish}, {Chong Wang}, {Jordan~L Boyd-Graber}, {and}
  {David~M Blei}. 2009.
\newblock \showarticletitle{Reading tea leaves: How humans interpret topic
  models}. In {\em Advances in neural information processing systems}.
  288--296.
\newblock


\bibitem{croskerry2003cognitive}
{Pat Croskerry}. 2003.
\newblock \showarticletitle{Cognitive forcing strategies in clinical
  decisionmaking}.
\newblock {\em Annals of emergency medicine\/} {41}, 1 (2003), 110--120.
\newblock


\bibitem{deslauriers2019measuring}
{Louis Deslauriers}, {Logan~S McCarty}, {Kelly Miller}, {Kristina Callaghan},
  {and} {Greg Kestin}. 2019.
\newblock \showarticletitle{Measuring actual learning versus feeling of
  learning in response to being actively engaged in the classroom}.
\newblock {\em Proceedings of the National Academy of Sciences\/} (2019),
  201821936.
\newblock


\bibitem{doshi2017towards}
{Finale Doshi-Velez} {and} {Been Kim}. 2017.
\newblock \showarticletitle{Towards a rigorous science of interpretable machine
  learning}.
\newblock {\em arXiv preprint arXiv:1702.08608\/} (2017).
\newblock


\bibitem{dzindolet2003role}
{Mary~T Dzindolet}, {Scott~A Peterson}, {Regina~A Pomranky}, {Linda~G Pierce},
  {and} {Hall~P Beck}. 2003.
\newblock \showarticletitle{The role of trust in automation reliance}.
\newblock {\em International journal of human-computer studies\/} {58}, 6
  (2003), 697--718.
\newblock


\bibitem{ericsson84:protocol}
{K~Anders Ericsson} {and} {Herbert~A Simon}. 1984.
\newblock {\em Protocol analysis: Verbal reports as data.}
\newblock the MIT Press.
\newblock


\bibitem{garbarino1997cognitive}
{Ellen~C Garbarino} {and} {Julie~A Edell}. 1997.
\newblock \showarticletitle{Cognitive effort, affect, and choice}.
\newblock {\em Journal of consumer research\/} {24}, 2 (1997), 147--158.
\newblock


\bibitem{garcia2018explainable}
{Francisco Javier~Chiyah Garcia}, {David~A Robb}, {Xingkun Liu}, {Atanas
  Laskov}, {Pedro Patron}, {and} {Helen Hastie}. 2018.
\newblock \showarticletitle{Explainable autonomy: A study of explanation styles
  for building clear mental models}. In {\em Proceedings of the 11th
  International Conference on Natural Language Generation}. 99--108.
\newblock


\bibitem{gilpin2018explaining}
{Leilani~H Gilpin}, {David Bau}, {Ben~Z Yuan}, {Ayesha Bajwa}, {Michael
  Specter}, {and} {Lalana Kagal}. 2018.
\newblock \showarticletitle{Explaining explanations: An overview of
  interpretability of machine learning}. In {\em 2018 IEEE 5th International
  Conference on data science and advanced analytics (DSAA)}. IEEE, 80--89.
\newblock


\bibitem{gorry1971framework}
{George~Anthony Gorry} {and} {Michael~S Scott~Morton}. 1971.
\newblock \showarticletitle{A framework for management information systems}.
\newblock  (1971).
\newblock


\bibitem{green2019disparate}
{Ben Green} {and} {Yiling Chen}. 2019a.
\newblock \showarticletitle{Disparate interactions: An algorithm-in-the-loop
  analysis of fairness in risk assessments}. In {\em Proceedings of the
  Conference on Fairness, Accountability, and Transparency}. ACM, 90--99.
\newblock


\bibitem{green2019principles}
{Ben Green} {and} {Yiling Chen}. 2019b.
\newblock \showarticletitle{The principles and limits of algorithm-in-the-loop
  decision making}.
\newblock {\em Proceedings of the ACM on Human-Computer Interaction\/} {3},
  CSCW (2019), 1--24.
\newblock


\bibitem{hauslschmid2017supportingtrust}
{Renate H{\"a}uslschmid}, {Max von Buelow}, {Bastian Pfleging}, {and} {Andreas
  Butz}. 2017.
\newblock \showarticletitle{Supportingtrust in autonomous driving}. In {\em
  Proceedings of the 22nd international conference on intelligent user
  interfaces}. ACM, 319--329.
\newblock


\bibitem{hoffman2018metrics}
{Robert~R Hoffman}, {Shane~T Mueller}, {Gary Klein}, {and} {Jordan Litman}.
  2018.
\newblock \showarticletitle{Metrics for explainable AI: Challenges and
  prospects}.
\newblock {\em arXiv preprint arXiv:1812.04608\/} (2018).
\newblock


\bibitem{johnston1994effects}
{Mary~E Johnston}, {Karl~B Langton}, {R~Brian Haynes}, {and} {Alix Mathieu}.
  1994.
\newblock \showarticletitle{Effects of computer-based clinical decision support
  systems on clinician performance and patient outcome: a critical appraisal of
  research}.
\newblock {\em Annals of internal medicine\/} {120}, 2 (1994), 135--142.
\newblock


\bibitem{kamar2016directions}
{Ece Kamar}. 2016.
\newblock \showarticletitle{Directions in Hybrid Intelligence: Complementing AI
  Systems with Human Intelligence.}. In {\em IJCAI}. 4070--4073.
\newblock


\bibitem{kamar2012combining}
{Ece Kamar}, {Severin Hacker}, {and} {Eric Horvitz}. 2012.
\newblock \showarticletitle{Combining human and machine intelligence in
  large-scale crowdsourcing}. In {\em Proceedings of the 11th International
  Conference on Autonomous Agents and Multiagent Systems-Volume 1}.
  International Foundation for Autonomous Agents and Multiagent Systems,
  467--474.
\newblock


\bibitem{kleinberg2017predictions}
{Jon Kleinberg}, {Himabindu Lakkaraju}, {Jure Leskovec}, {Jens Ludwig}, {and}
  {Sendhil Mullainathan}. 2017.
\newblock \showarticletitle{{Human Decisions and Machine Predictions*}}.
\newblock {\em The Quarterly Journal of Economics\/} {133}, 1 (08 2017),
  237--293.
\newblock
\showISSN{0033-5533}
\showDOI{%
\url{http://dx.doi.org/10.1093/qje/qjx032}}


\bibitem{kool18:mental}
{Wouter Kool} {and} {Matthew Botvinick}. 2018.
\newblock \showarticletitle{Mental labour}.
\newblock {\em Nature human behaviour\/} {2}, 12 (2018), 899--908.
\newblock


\bibitem{kool2010decision}
{Wouter Kool}, {Joseph~T McGuire}, {Zev~B Rosen}, {and} {Matthew~M Botvinick}.
  2010.
\newblock \showarticletitle{Decision making and the avoidance of cognitive
  demand.}
\newblock {\em Journal of Experimental Psychology: General\/} {139}, 4 (2010),
  665.
\newblock


\bibitem{kulesza2015principles}
{Todd Kulesza}, {Margaret Burnett}, {Weng-Keen Wong}, {and} {Simone Stumpf}.
  2015.
\newblock \showarticletitle{Principles of explanatory debugging to personalize
  interactive machine learning}. In {\em Proceedings of the 20th international
  conference on intelligent user interfaces}. ACM, 126--137.
\newblock


\bibitem{lage2019human}
{Isaac Lage}, {Emily Chen}, {Jeffrey He}, {Menaka Narayanan}, {Been Kim},
  {Samuel~J Gershman}, {and} {Finale Doshi-Velez}. 2019.
\newblock \showarticletitle{Human Evaluation of Models Built for
  Interpretability}. In {\em Proceedings of the AAAI Conference on Human
  Computation and Crowdsourcing}, Vol.~7. 59--67.
\newblock


\bibitem{lai2019human}
{Vivian Lai} {and} {Chenhao Tan}. 2019.
\newblock \showarticletitle{On human predictions with explanations and
  predictions of machine learning models: A case study on deception detection}.
  In {\em Proceedings of the Conference on Fairness, Accountability, and
  Transparency}. 29--38.
\newblock


\bibitem{lakkaraju2016interpretable}
{Himabindu Lakkaraju}, {Stephen~H Bach}, {and} {Jure Leskovec}. 2016.
\newblock \showarticletitle{Interpretable decision sets: A joint framework for
  description and prediction}. In {\em Proceedings of the 22nd ACM SIGKDD
  international conference on knowledge discovery and data mining}. ACM,
  1675--1684.
\newblock


\bibitem{lakkaraju2019fool}
{Himabindu Lakkaraju} {and} {Osbert Bastani}. 2019.
\newblock \showarticletitle{"How do I fool you?": Manipulating User Trust via
  Misleading Black Box Explanations}.
\newblock {\em arXiv preprint arXiv:1911.06473\/} (2019).
\newblock


\bibitem{lambe16:dual}
{Kathryn~Ann Lambe}, {Gary O'Reilly}, {Brendan~D. Kelly}, {and} {Sarah
  Curristan}. 2016.
\newblock \showarticletitle{{Dual-process cognitive interventions to enhance
  diagnostic reasoning: A systematic review}}.
\newblock {\em BMJ Quality and Safety\/} {25}, 10 (2016), 808--820.
\newblock
\showISSN{20445415}
\showDOI{%
\url{http://dx.doi.org/10.1136/bmjqs-2015-004417}}


\bibitem{lee2004trust}
{John~D Lee} {and} {Katrina~A See}. 2004.
\newblock \showarticletitle{Trust in automation: Designing for appropriate
  reliance}.
\newblock {\em Human factors\/} {46}, 1 (2004), 50--80.
\newblock


\bibitem{muir1987trust}
{Bonnie~M Muir}. 1987.
\newblock \showarticletitle{Trust between humans and machines, and the design
  of decision aids}.
\newblock {\em International journal of man-machine studies\/} {27}, 5-6
  (1987), 527--539.
\newblock


\bibitem{poursabzi2018manipulating}
{Forough Poursabzi-Sangdeh}, {Daniel~G Goldstein}, {Jake~M Hofman},
  {Jennifer~Wortman Vaughan}, {and} {Hanna Wallach}. 2018.
\newblock \showarticletitle{Manipulating and measuring model interpretability}.
\newblock {\em arXiv preprint arXiv:1802.07810\/} (2018).
\newblock


\bibitem{ribeiro2016should}
{Marco~Tulio Ribeiro}, {Sameer Singh}, {and} {Carlos Guestrin}. 2016.
\newblock \showarticletitle{Why should I trust you?: Explaining the predictions
  of any classifier}. In {\em Proceedings of the 22nd ACM SIGKDD international
  conference on knowledge discovery and data mining}. ACM, 1135--1144.
\newblock


\bibitem{selvaraju2017grad}
{Ramprasaath~R Selvaraju}, {Michael Cogswell}, {Abhishek Das}, {Ramakrishna
  Vedantam}, {Devi Parikh}, {and} {Dhruv Batra}. 2017.
\newblock \showarticletitle{Grad-cam: Visual explanations from deep networks
  via gradient-based localization}. In {\em Proceedings of the IEEE
  International Conference on Computer Vision}. 618--626.
\newblock


\bibitem{shah2008heuristics}
{Anuj~K Shah} {and} {Daniel~M Oppenheimer}. 2008.
\newblock \showarticletitle{Heuristics made easy: An effort-reduction
  framework.}
\newblock {\em Psychological bulletin\/} {134}, 2 (2008), 207.
\newblock


\bibitem{sherbino14:ineffectiveness}
{Jonathan Sherbino}, {Kulamakan Kulasegaram}, {Elizabeth Howey}, {and}
  {Geoffrey Norman}. 2014.
\newblock \showarticletitle{{Ineffectiveness of cognitive forcing strategies to
  reduce biases in diagnostic reasoning: A controlled trial}}.
\newblock {\em Canadian Journal of Emergency Medicine\/} {16}, 1 (2014),
  34--40.
\newblock
\showISSN{14818035}
\showDOI{%
\url{http://dx.doi.org/10.2310/8000.2013.130860}}


\bibitem{inductive-analysis}
{David~R. Thomas}. 2006.
\newblock \showarticletitle{A General Inductive Approach for Analyzing
  Qualitative Evaluation Data}.
\newblock {\em American Journal of Evaluation\/} {27}, 2 (2006), 237--246.
\newblock
\showDOI{%
\url{http://dx.doi.org/10.1177/1098214005283748}}


\bibitem{tversky1974judgment}
{Amos Tversky} {and} {Daniel Kahneman}. 1974.
\newblock \showarticletitle{Judgment under uncertainty: Heuristics and biases}.
\newblock {\em science\/} {185}, 4157 (1974), 1124--1131.
\newblock


\bibitem{wang2019designing}
{Danding Wang}, {Qian Yang}, {Ashraf Abdul}, {and} {Brian~Y Lim}. 2019.
\newblock \showarticletitle{Designing Theory-Driven User-Centric Explainable
  AI}. In {\em Proceedings of the 2019 CHI Conference on Human Factors in
  Computing Systems}. ACM, 601.
\newblock


\bibitem{wang2007theoretical}
{Yingxu Wang}. 2007.
\newblock \showarticletitle{The theoretical framework of cognitive
  informatics}.
\newblock {\em International Journal of Cognitive Informatics and Natural
  Intelligence (IJCINI)\/} {1}, 1 (2007), 1--27.
\newblock


\bibitem{wang2006layered}
{Yingxu Wang}, {Ying Wang}, {Shushma Patel}, {and} {Dilip Patel}. 2006.
\newblock \showarticletitle{A layered reference model of the brain (LRMB)}.
\newblock {\em IEEE Transactions on Systems, Man, and Cybernetics, Part C
  (Applications and Reviews)\/} {36}, 2 (2006), 124--133.
\newblock


\bibitem{weitz2019you}
{Katharina Weitz}, {Dominik Schiller}, {Ruben Schlagowski}, {Tobias Huber},
  {and} {Elisabeth Andr{\'e}}. 2019.
\newblock \showarticletitle{Do you trust me?: Increasing User-Trust by
  Integrating Virtual Agents in Explainable AI Interaction Design}. In {\em
  Proceedings of the 19th ACM International Conference on Intelligent Virtual
  Agents}. ACM, 7--9.
\newblock


\bibitem{wilson2001encyclopedia}
{Robert~Andrew Wilson} {and} {Frank~C Keil}. 2001.
\newblock {\em The MIT encyclopedia of the cognitive sciences}.
\newblock


\bibitem{yin2019understanding}
{Ming Yin}, {Jennifer Wortman~Vaughan}, {and} {Hanna Wallach}. 2019.
\newblock \showarticletitle{Understanding the Effect of Accuracy on Trust in
  Machine Learning Models}. In {\em Proceedings of the 2019 CHI Conference on
  Human Factors in Computing Systems}. ACM, 279.
\newblock


\bibitem{zeleznikow2004building}
{John Zeleznikow}. 2004.
\newblock \showarticletitle{Building intelligent legal decision support
  systems: Past practice and future challenges}.
\newblock In {\em Applied Intelligent Systems}. Springer, 201--254.
\newblock


\bibitem{Zhou2018InterpretableBD}
{Bolei Zhou}, {Yiyou Sun}, {David Bau}, {and} {Antonio Torralba}. 2018.
\newblock \showarticletitle{Interpretable Basis Decomposition for Visual
  Explanation}. In {\em ECCV}. 119--134.
\newblock


\end{thebibliography}
\bibliographystyle{SIGCHI-Reference-Format}

\end{document}